\documentclass[sigconf]{acmart}

\usepackage{placeins}
\usepackage{hyperref}
\usepackage{subfig}
\usepackage{multirow}
\usepackage{tikz}
\usepackage{svg}
\usepackage{todonotes}
\usepackage{multicol}
\def\BibTeX{{\rm B\kern-.05em{\sc i\kern-.025em b}\kern-.08emT\kern-.1667em\lower.7ex\hbox{E}\kern-.125emX}}
    
\copyrightyear{2019}
\acmYear{2019}
\setcopyright{acmlicensed}
\acmConference[]{ }{}{}
\acmBooktitle{}
\acmPrice{15.00} 
\acmDOI{10.1145/1122445.1122456} 
\acmISBN{978-1-4503-9999-9/18/06} 
\renewcommand\footnotetextcopyrightpermission[1]{} 

\begin{document}

%
\title{Multimodal Classification of Urban Micro-Events}

%
\author{Maarten Sukel}
\affiliation{%
  \institution{University of Amsterdam}
  \city{Amsterdam}
  \country{The Netherlands}}
\email{m.m.sukel@uva.nl}

\author{Stevan Rudinac}
\affiliation{%
  \institution{University of Amsterdam}
  \city{Amsterdam}
  \country{The Netherlands}}
\email{s.rudinac@uva.nl}

\author{Marcel Worring}
\affiliation{%
  \institution{University of Amsterdam}
  \city{Amsterdam}
  \country{The Netherlands}}
\email{m.worring@uva.nl}

%
\renewcommand{\shortauthors}{Sukel, Rudinac and Worring}

\newcommand{\ms}[1]{\todo[inline,color=yellow!40,author=Maarten]{ #1}}
\newcommand{\sr}[1]{\todo[inline,color=magenta!40,author=Stevan]{ #1}}
\newcommand{\mw}[1]{\todo[inline,color=red!40,author=Marcel]{ #1}}

%
\begin{abstract}
In this paper we seek methods to effectively detect urban micro-events. Urban micro-events are events which occur in cities, have limited geographical coverage and typically affect only a small group of citizens. Because of their scale these are difficult to identify in most data sources. However, by using citizen sensing to gather data, detecting them becomes feasible. The data gathered by citizen sensing is often multimodal and, as a consequence, the information required to detect urban micro-events is distributed over multiple modalities. This makes it essential to have a classifier capable of combining them. In this paper we explore several methods of creating such a classifier, including early, late, hybrid fusion and representation learning using multimodal graphs. We evaluate performance on a real world dataset obtained from a live citizen reporting system. We show that a multimodal approach yields higher performance than unimodal alternatives. Furthermore, we demonstrate that our hybrid combination of early and late fusion with multimodal embeddings performs best in classification of urban micro-events.
\end{abstract}

%
%



%


%

\maketitle
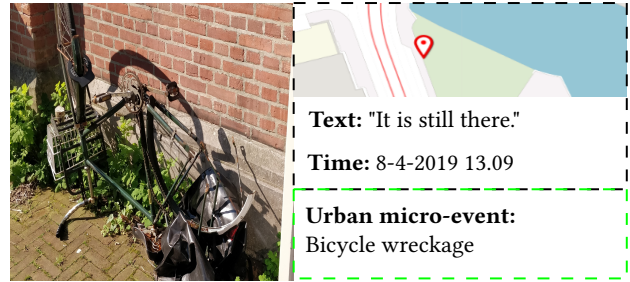
\begin{figure}

\tikzset{every picture/.style={line width=0.75pt}} 

\begin{tikzpicture}[x=0.75pt,y=0.75pt,yscale=-1,xscale=1]

\draw (89.5,79) node  {\includegraphics[width=107.25pt,height=105pt]{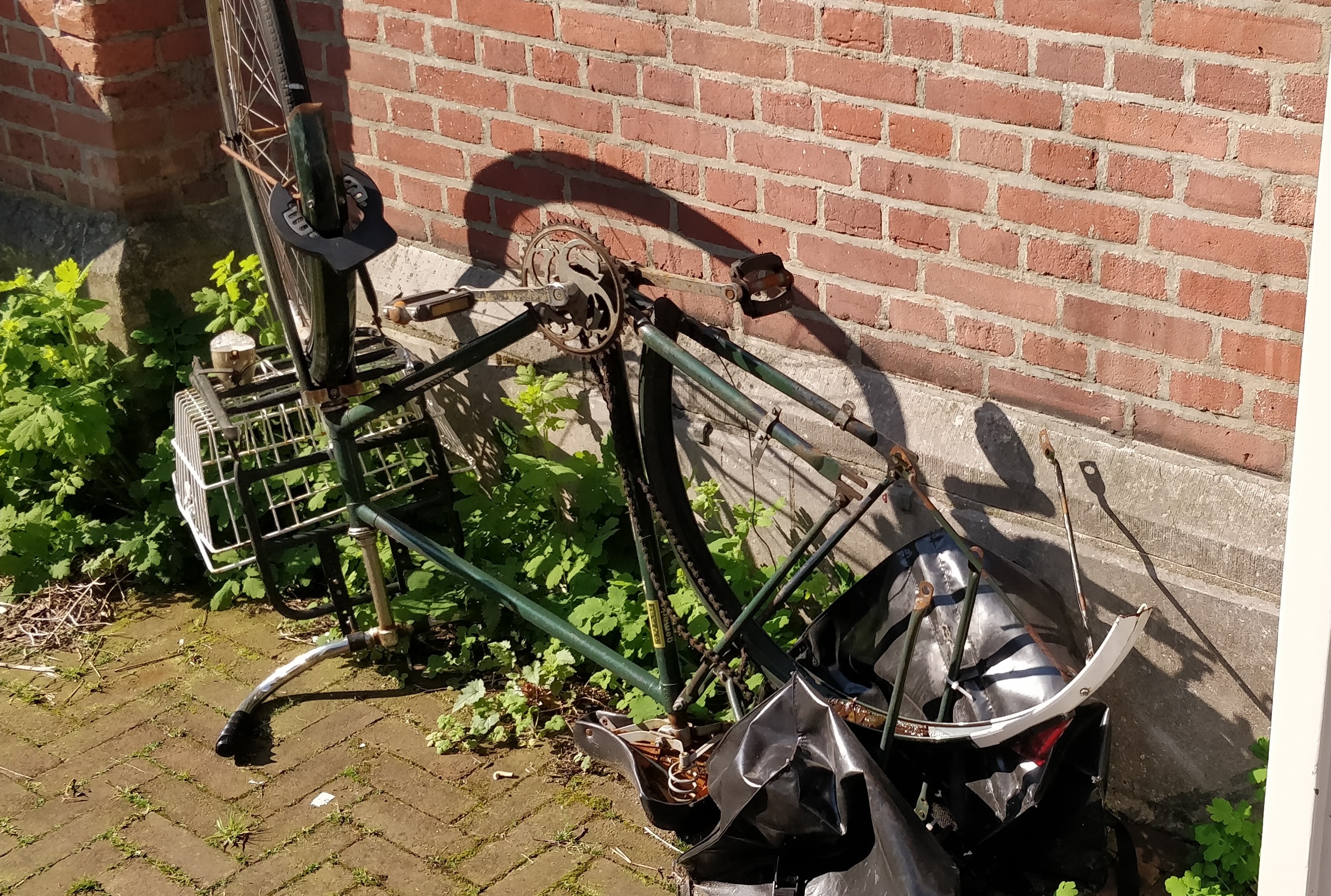}};
\draw (245,32.5) node  {\includegraphics[width=126pt,height=35.25pt]{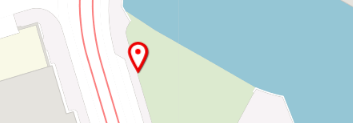}};
\draw  [dash pattern={on 4.5pt off 4.5pt}] (161,9) -- (329,9) -- (329,103) -- (161,103) -- cycle ;
\draw  [color={rgb, 255:red, 0; green, 255; blue, 0 }  ,draw opacity=1 ][dash pattern={on 4.5pt off 4.5pt}] (161,103) -- (329,103) -- (329,148) -- (161,148) -- cycle ;

\draw (222,67.5) node  [align=left] {\textbf{Text:} "It is still there."};
\draw (220,89.5) node  [align=left] {\textbf{Time:} 8-4-2019 13.09};
\draw (220,124.5) node  [align=left] {\textbf{Urban micro-event: }\\Bicycle wreckage};

\end{tikzpicture}

  \caption{Example of how an urban micro-event has to be detected between multiple modalities.}

  \label{fig:teaser}
\end{figure}

\section{Introduction}
\label{section:introduction}
Cities are living organisms where numerous events take place at different geographical and temporal scales. Some of these events are macroscopic, involving a large geographical area and a large number of people. Other events occur in a very limited geographical area and have a smaller number of people involved. In this paper we focus on such small scale events, referring to them as the \textit{urban micro-events}. In particular we focus on the subclass of urban micro-events that lead to a service request. Examples of such urban micro-events are speeding boats, graffiti on walls, trash on the street, broken streetlights or a bicycle wreckage that needs to be removed (cf. Figure~\ref{fig:teaser}). Urban micro-events are difficult to find due to their small scale and the large variety of forms they may take. However since the citizens are often motivated to help increase city livability, they report possible issues, creating valuable data sources for the detection of these urban micro-events. Urban micro-events are likely to cause issues resulting in a negative impact on livability and thus citizens in the neighborhood. In this paper we focus on detecting those kind of urban micro-events.



\begin{figure*}[h!]
\centering
    \includegraphics[height=1\textwidth,angle=270]{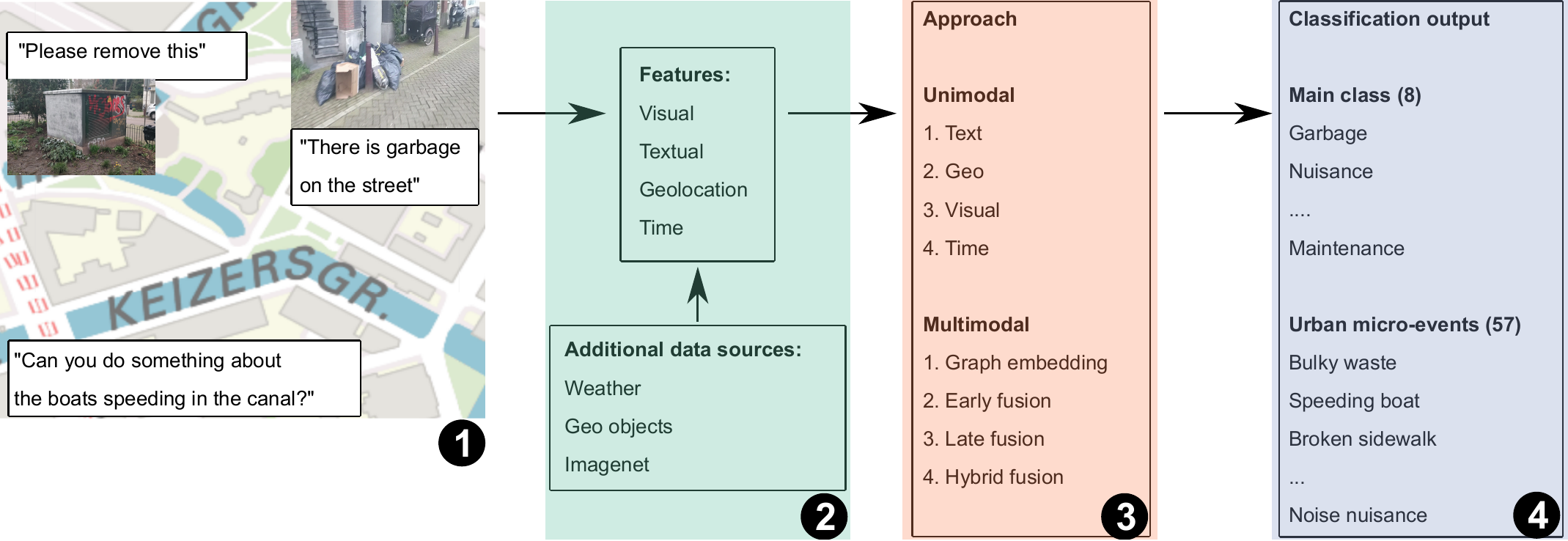}
   \caption{To detect urban micro-events in data gathered by using citizen sensing (1), features are extracted and additional contextual data is added (2), unimodal and multimodal classifiers are created (3) and performance is evaluated on a real world dataset (4).}
   \label{fig:schema_graph}
\end{figure*}

Different types of events have been studied in various areas of multimedia. For example, events have been intensively studied in video analysis \cite{2018trecvidawad}, but such events normally have a clear and consistent visual pattern and occur over multiple frames. The task we try to perform is different in nature because the image or text might not directly mention the related issue. In recent years research has been done on combining social multimedia with satellite imagery to detect natural disaster related events \cite{bischke2017multimedia} and detecting if tweets related to real-world events are fake or real \cite{boididou2014challenges}. These tasks combine multimedia data to perform classification, which makes them relevant to our task at hand. However, the target events are very different from the urban micro-events we attempt to classify.
For detecting urban micro-events the above methods give inspiration, but are not directly applicable.

Using citizens to collect data about their surroundings is often called the \textit{citizen as a sensor} paradigm \cite{boulos2011crowdsourcing} \cite{tang2016research}, and alternatively, participatory sensing or human-in-the-loop sensing. 
Such a paradigm can be also deployed to collect, analyze and mine information about events. Pervasiveness of smart phones with inbuilt high-quality sensors makes such collected information increasingly valuable. The data used for experiments in this paper has been collected by a semi structured citizen as a sensor system, in which citizens report on what they deem relevant, without being restricted by structured forms.

An application of the citizen as a sensor paradigm frequently seen in cities around the world is a service where residents can report issues in the public space, which creates data containing rich information about urban micro-events. For example, in most cities in the United States it is possible to call 311 for non-emergency service requests, report issues through web forms or contact the local government through social media channels such as Twitter or WhatsApp. These request often report and describe an urban micro-event. Doing so, cities are trying to get closer to the citizens by increasingly responding to urban service requests, which results in several difficulties. 
First, the type of issue needs to be precisely determined in order to offer timely and appropriate solution. 

The classification of the reports is a challenging task due to the heterogeneous nature of urban micro-events, which range from anything related to nuisance in the public area caused by begging or boats with loud music to potholes and dangerous traffic situations (cf. Figure~\ref{fig:schema_graph}). Since a large number of cities have a citizen report system that could benefit from the accurate classification of urban micro-events, effective solution to the problem could have a large potential for improving city livability.


The reports made by citizens often consist of text, image, spatial and temporal data and require the selection of an issue category. Since the citizen is not always familiar with the meaning of the class, this creates several problems. One of the main problems is that choosing the wrong class results in the issue not being sent to the right department, which in turn results in the issue not being solved or being solved with a delay. A solution for this is letting experts decide the correct class. However this is a labor intensive exercise which is also likely to cause delays. 
The use of an automated classifier that performs better than both citizen and the expert would allow for a faster detection of urban micro-events and their underlying issues, and consequently lead to a timely issue resolution and a reduced negative impact on the city livability. 
Since more and more service requests have relevant information in several different modalities, it is essential to be able to combine them. An example is a combination of text that reports trash on the street, and the associated image that shows the type of trash. Complete understanding of the underlying issue thus requires information extracted from both text and visual content. Another example is a noise complaint, where the time and location of the report could be relevant for how the issue should be classified. The classifier in this case has to be capable of effectively extracting relevant features out of a wide range of data types, including visual, textual, spatial and temporal data. 

For creating a multi-modal classifier we consider several fusion schemes: Early, late and hybrid fusion. In early fusion the features are first extracted from each modality and then combined in a joint representation. The approaches to early fusion range from simple concatenation to complex graph embeddings. Late fusion, on the other hand, combines different modalities by first performing classification on the unimodal feature vectors, and then using such obtained classification results as an input into the classifier combiner (i.e. meta-estimator). Both classic \cite{kittler1996combining} and more recent studies \cite{mohandes2018classifiers} show that the optimal choice of classifier combining technique depends on the application. For example, research in video retrieval shows that late fusion tends to give slightly better performance, but where early fusion performs better the difference is more significant \cite{snoek2005early}. Since we are combining four modalities a hybrid fusion technique combining early and late fusion could result in even better performance. 


To arrive at an optimal method, in this paper we utilize a wide range of modalities and modality fusion techniques for an accurate classification of urban micro-events. The contributions of this paper can be summarized as follows:

\begin{itemize} 
 \item We present several fusion methods, both early, late and hybrid, to create a multi-modal classifier capable of detecting urban micro-events that uses textual, visual, spatial and temporal data.
 \item We evaluate the resulting multi-modal learning method on a real life system deployed in a major city for detecting urban micro-events.
 \item We elaborate on the usefulness of textual, visual, spatial and temporal features for the classification of urban micro-events by doing an extended evaluation.

\end{itemize}

\pagebreak
\section{Related work}
\label{section:related_work}

In this section we discuss related work. First we start with approaches to multimodal classification, from there we explore the citizen as a sensor paradigm and customer feedback systems. Finally we discuss approaches for event detection.

\subsection{Multimodal classification}

Multimodal classification has been a long-standing research topic in the multimedia community and over time a number of excellent solutions have been proposed.
There are several general approaches, the most common being early and late fusion. Below we survey recent trends in both ``schools''.

Recent research efforts revolve around creating a joint item representation from the features of different modalities. For example, information extracted from the visual content and the text could be combined by generating a new `imagined' vector as in \cite{collell2017imagined} or by using a common subspace as in \cite{wang2017adversarial}. This is, however, not true for the data set we are using since the image labels are often more subjective. For example a picture of noise disturbance can be many different things each with their own visual appearance. For the same reasons, using the text to determine attention as in \cite{li2018read} is not likely to work on urban micro-events. However, as shown in \cite{zhang2017improving}, using visual content to resolve ambiguities can significantly improve upon text-based event extraction results, even when the modalities are not well aligned.

Multimodal classification requires extracting features from the visual content. Features can be extracted from images by using e.g. convolutional neural networks. These features can be visual concepts like a sheep or a bicycle, but often also more abstract visual features like the vector of elements defining the layer before the softmax as in \cite{collell2017imagined}. A group of deep networks, referred to as residual networks or ResNet \cite{he2016deep}, achieved particularly good performance in various task such as image recognition or object detection.
Such extracted features generally yield better results than the traditional alternatives, such as Bag of visual words, SIFT or HOG \cite{kiela2014learning,li2016event}, which is why we consider them a promising starting point for building multimodal representation of citizen reports. 

Graphs have also been deployed for embedding different modalities beyond text and visual content, and then the affinities between multimedia items were computed using e.g. random walks with restarts as in \cite{rudinac2013generating}. With the advent of deep learning, the general idea was revived in the approaches such as DeepWalk \cite{Perozzi:2014:DOL:2623330.2623732}. More recently, Grover and Leskovec proposed a node2vec framework for learning a low-dimensional representations of the nodes in a graph by optimizing a neighborhood preserving objective, allowing representation of complex networks \cite{GroverL16}. We conjecture that this approach may be effective in encoding complex relations between different modalities in our collection, ranging from text and visual content to geolocation, time and weather conditions. 


\subsection{Citizen as a sensor} 
The use of citizen as a sensor brings multiple challenges \cite{boulos2011crowdsourcing}. Typically, domain experts are tasked with determining the category of the issue described by the report, but their time is limited and costly.
The range of reported issues is also broad, requiring knowledge of multiple domains and location specific information. Another challenge is that citizens have no way of discerning truth from falsehood. The same applies for analyzing the reports, citizens that are interested in skewing information can create bias in the data. The use of machine learning for processing reports also has shortcomings since it requires training data and feedback. However, having humans perform the classification is resource intensive task and, in addition, they may become tired over time, possibly missing or misinterpreting signals. 

Customer feedback data is used for the creation of classifiers: Textual classification of customer feedback chats is discussed in \cite{Park2015mining}. Being able to predict customer satisfaction based on textual data woud certainly be a nice addition to our approach, but our primary goal is detection of micro events in an urban environment. In \cite{lin2017sentinlp} methods to detect classes (i.e. comment, request, bug, complaint, meaningless, and undetermined) in customer service data are compared and the best scoring method is a bidirectional LSTM+CNN. Similarly, \cite{liu2018understanding} explores understanding of the same classes in multilingual customer feedback. The data used has a heterogeneous nature, consisting of multiple languages. These categories are different from the urban micro-events we seek to classify, however the type of textual input data and output have a similar nature so the best scoring methods might also apply for the classification of urban micro events. 

Several designs of citizen feedback systems are compared in \cite{offenhuber2014infrastructure} and several important factors of how such a system impacts the interaction between citizens and municipalities are discussed. The study identifies the way in which category selection works as one of the most important factors determining effectiveness of the system.   

Finally, as stressed by Tang et al., utilizing the full potential of citizen reports requires effective methods for their analysis and categorization, based on increasingly heterogeneous multimedia data they contain \cite{Tang:2016:RCD:2964284.2976761}. In this paper we present several solutions to the category selection problem. 

\subsection{Event detection}

Event detection has been a long-standing interest of the research community. Examples are plentiful and range from classic works on detection of events in news articles using text retrieval and clustering techniques \cite{yang1998study} to multimedia event detection in video materials \cite{2018trecvidawad}. Methods for finding events in news and social data streams are particularly well researched \cite{chen2017fine,huang2018spatial,qian2016multi,yang2016abnormal,walther2013geo}, however the scope, such as festivals and international incidents, is of a different geographical scale than the urban micro-events we attempt to classify. Still the fine-grained characterization of events from social data streams might prove a useful technique for our purpose.  Social data is also used to detect emergency situations \cite{kanojia2016civique} and combined with satellite images to detect natural disasters  \cite{bischke2017multimedia}. Social data is also used to determine whether real-world events are fake or real, by fusing multiple modalities (social, text and visual) \cite{boididou2014challenges,jin2017multimodal}.

When working with social data it is possible to use additional information such as URLs, users and hash tags as in \cite{chen2016context} and \cite{mehrotra2013improving}. For example, \cite{mehrotra2013improving} utilize hashtags and user handles for pooling tweets and extracting latent topics of a higher quality. Similarly, in \cite{hu2018decode} textual and geographical data are combined to detect different type of users in geographical and textual social media data. Due to privacy and ethical concerns, no information about users is collected by the citizen reporting system discussed in this paper. However when using metadata, such as time and geolocation, inspiration can be drawn from the above-mentioned related work.  Methods for detecting generic micro events using multi-modal techniques are described in \cite{jayarajah2016can}. These micro events are different from the urban micro-events we seek. They are defined as transient occurrences that occur within larger events. 

Most related work in event detection seeks for events of a different nature, having a different geographical coverage and affecting a different number of citizens. However, the additional use of multimedia content might prove useful since we also try to create a multi-modal classifier. While most approaches on event detection center on social media data or news articles, the main focus of this paper is detection of urban micro-events in a real world data set containing extremely heterogeneous multimedia data. Our objective is therefore aligned with the recent efforts of multimedia community towards rethinking the very concept of event in the age of multimedia data that goes beyond simple text and visual modalities \cite{rudinac2018rethinking}.   

\section{Approach}
In this section we describe our approach to classifying urban micro-events, which consists of the following steps: (3.1)  Discovering features for different modalities, (3.2) Creating unimodal classifiers using these features (3.3) Creating multimodal classifiers combining these modalities.

\subsection{Features}
\label{section:features}
We propose several methods for extracting features out of text, image, geo and temporal data. The source code that shows how the features are extracted on a sample dataset can be found in the linked repository \cite{github2019}.

\subsubsection{Textual}

Since the textual description of an urban micro-event often has a large quantity of information, we will explore several methods of representing the data and evaluate what method works best for detecting urban micro-events. Motivated by the recent research in information retrieval community, which shows that for some categories word embeddings work better, while for the others traditional vector space models still yield a better performance \cite{VanGysel:2016:UES:2872427.2882974}, we decided to evaluate both TF-IDF \cite{ramos2003using} and word2vec \cite{mikolov2013efficient} representations. Our initial experiments with different variants of Latent Dirichlet Allocation \cite{Blei:2012:PTM:2133806.2133826} yielded unsatisfying performance, mostly due to a short length of the reports and a varying quality of conversational language used in them, which is why we do not report on them in this paper.

\subsubsection{Visual} 

Similar to \cite{kiela2014learning}, as the visual features we will use the 2048-dimensional output of the last layer before softmax of the ResNet50 model \cite{he2016deep}, pre-trained on ImageNet \cite{deng2009imagenet}. 
Another possibility is using output of the network containing confidences for ImageNet classes, but our visual data is too different from ImageNet, which might results in poor classification performance. For example, bicycles are likely correctly detected, unlike the garbage containers and garbage bags. By using the hidden representation the features are more general. For the creation of the graph embedding the output of the network will be used to reduce the number of nodes.

\subsubsection{Geo objects}
For creating a fingerprint capturing the geographical location of reports, a reference database with geographical data that describes the environment will be used. Using this we can create new features such as the distance to the closest container, the mean average distance of five double flowered chestnut trees and the number of residential buildings within $n$ meters. In our case the data about geo objects and their location consist of 552.999 geo objects extracted from \url{https://maps.amsterdam.nl/}. This results in 1856 features describing the proximity and density of objects in the environment which are defined as follows 

\begin{itemize}
    \item Proximity is represented by taking the distance to the closest geo object of each type, but also by taking the mean of the closest five, ten and hundred objects per type.
    \item Density is represented by counting the occurrences of a geo object type within 25, 50, 100 and 200 meter.
\end{itemize}

Using available historical geographical data of urban micro-events we will create a historical profile of the area by using the same feature creation method as for the geographical objects, creating another 472 geo features on 57 different types of urban micro-events.

\begin{figure}
\centering
\subfloat[Proximity]{{\includegraphics[width=0.20\textwidth,height=0.20\textwidth]{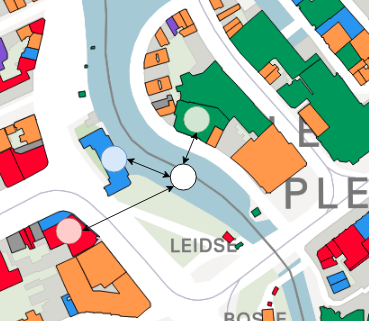} }}%
\qquad
\subfloat[Density]{{\includegraphics[width=0.20\textwidth,height=0.20\textwidth]{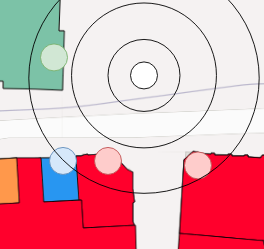} }}%
\caption{Examples of how the density and proximity of geographical object types is captured. }%
\label{fig:example}%
\end{figure}

\subsubsection{Temporal and weather features}
Using one-hot encoded temporal data of month, weekday and hour of day, and historical weather data per hour such as temperature, wind speed, snow and rain we will create 37 time features and 18 weather features.

\subsection{Unimodal classifiers}
\label{section:unimodal}

\begin{figure}
\tikzset{every picture/.style={line width=0.75pt}} 

\begin{tikzpicture}[x=0.75pt,y=0.75pt,yscale=-0.65,xscale=0.65]


\draw    (224,97.15) .. controls (244.17,119.15) and (229.17,141.15) .. (252,150) ;

\draw    (302,150) .. controls (315.17,131.15) and (317.17,122.15) .. (311,83) ;

\draw    (311,83) .. controls (331.17,109.15) and (329.17,143.15) .. (365,150) ;

\draw    (456,82) .. controls (466.17,99.15) and (441.17,148.15) .. (415,150) ;

\draw    (475,190) .. controls (456.17,179.15) and (427.17,183.15) .. (415,150) ;

\draw    (277,175) .. controls (300.17,190.15) and (305.17,214.15) .. (293.17,217.15) ;

\draw    (293.17,217.15) .. controls (333.17,187.15) and (350,205) .. (390,175) ;

\draw    (390,175) .. controls (415.17,184.15) and (423.17,207.15) .. (429.17,209.15) ;

\draw    (190.17,137.15) .. controls (225.17,123.15) and (212,180) .. (252,150) ;

\draw    (229.17,178.15) .. controls (239.17,164.15) and (223.17,183.15) .. (252,150) ;

\draw    (242,71) .. controls (279.17,56.15) and (325.17,182.15) .. (365,150) ;

\draw    (376.17,222.15) .. controls (416.17,192.15) and (374.17,201.15) .. (390,175) ;

\draw  [dash pattern={on 0.84pt off 2.51pt}]  (139.17,138.15) .. controls (127.17,174.15) and (225.17,357.15) .. (443,258) ;

\draw  [color={rgb, 255:red, 184; green, 233; blue, 134 }  ,draw opacity=1 ][line width=3]  (129,52) .. controls (129,38.19) and (140.19,27) .. (154,27) .. controls (167.81,27) and (179,38.19) .. (179,52) .. controls (179,65.81) and (167.81,77) .. (154,77) .. controls (140.19,77) and (129,65.81) .. (129,52) -- cycle ;
\draw    (168.17,75.15) .. controls (193.17,77.15) and (210.17,149.15) .. (252,150) ;

\draw    (390,125) .. controls (391.17,105.15) and (379.17,121.15) .. (383,89) ;

\draw  [dash pattern={on 0.84pt off 2.51pt}]  (173.17,35.15) .. controls (213.17,5.15) and (347.17,8.15) .. (383,39) ;

\draw  [color={rgb, 255:red, 126; green, 211; blue, 33 }  ,draw opacity=1 ][line width=3.75]  (192,71) .. controls (192,57.19) and (203.19,46) .. (217,46) .. controls (230.81,46) and (242,57.19) .. (242,71) .. controls (242,84.81) and (230.81,96) .. (217,96) .. controls (203.19,96) and (192,84.81) .. (192,71) -- cycle ;
\draw  [color={rgb, 255:red, 74; green, 144; blue, 226 }  ,draw opacity=1 ][line width=3.75]  (286,58) .. controls (286,44.19) and (297.19,33) .. (311,33) .. controls (324.81,33) and (336,44.19) .. (336,58) .. controls (336,71.81) and (324.81,83) .. (311,83) .. controls (297.19,83) and (286,71.81) .. (286,58) -- cycle ;
\draw  [color={rgb, 255:red, 208; green, 2; blue, 27 }  ,draw opacity=1 ][line width=3.75]  (365,150) .. controls (365,136.19) and (376.19,125) .. (390,125) .. controls (403.81,125) and (415,136.19) .. (415,150) .. controls (415,163.81) and (403.81,175) .. (390,175) .. controls (376.19,175) and (365,163.81) .. (365,150) -- cycle ;
\draw  [color={rgb, 255:red, 208; green, 2; blue, 27 }  ,draw opacity=1 ][line width=3.75]  (252,150) .. controls (252,136.19) and (263.19,125) .. (277,125) .. controls (290.81,125) and (302,136.19) .. (302,150) .. controls (302,163.81) and (290.81,175) .. (277,175) .. controls (263.19,175) and (252,163.81) .. (252,150) -- cycle ;
\draw  [color={rgb, 255:red, 80; green, 227; blue, 194 }  ,draw opacity=1 ][line width=3.75]  (140,129) .. controls (140,115.19) and (151.19,104) .. (165,104) .. controls (178.81,104) and (190,115.19) .. (190,129) .. controls (190,142.81) and (178.81,154) .. (165,154) .. controls (151.19,154) and (140,142.81) .. (140,129) -- cycle ;
\draw  [color={rgb, 255:red, 65; green, 117; blue, 5 }  ,draw opacity=1 ][line width=3.75]  (183,191) .. controls (183,177.19) and (194.19,166) .. (208,166) .. controls (221.81,166) and (233,177.19) .. (233,191) .. controls (233,204.81) and (221.81,216) .. (208,216) .. controls (194.19,216) and (183,204.81) .. (183,191) -- cycle ;
\draw  [color={rgb, 255:red, 144; green, 19; blue, 254 }  ,draw opacity=1 ][line width=3.75]  (244,231) .. controls (244,217.19) and (255.19,206) .. (269,206) .. controls (282.81,206) and (294,217.19) .. (294,231) .. controls (294,244.81) and (282.81,256) .. (269,256) .. controls (255.19,256) and (244,244.81) .. (244,231) -- cycle ;
\draw  [color={rgb, 255:red, 65; green, 117; blue, 5 }  ,draw opacity=1 ][line width=3]  (335,240) .. controls (335,226.19) and (346.19,215) .. (360,215) .. controls (373.81,215) and (385,226.19) .. (385,240) .. controls (385,253.81) and (373.81,265) .. (360,265) .. controls (346.19,265) and (335,253.81) .. (335,240) -- cycle ;
\draw  [color={rgb, 255:red, 126; green, 211; blue, 33 }  ,draw opacity=1 ][line width=3.75]  (475,190) .. controls (475,176.19) and (486.19,165) .. (500,165) .. controls (513.81,165) and (525,176.19) .. (525,190) .. controls (525,203.81) and (513.81,215) .. (500,215) .. controls (486.19,215) and (475,203.81) .. (475,190) -- cycle ;
\draw  [color={rgb, 255:red, 74; green, 144; blue, 226 }  ,draw opacity=1 ][line width=3.75]  (431,57) .. controls (431,43.19) and (442.19,32) .. (456,32) .. controls (469.81,32) and (481,43.19) .. (481,57) .. controls (481,70.81) and (469.81,82) .. (456,82) .. controls (442.19,82) and (431,70.81) .. (431,57) -- cycle ;
\draw  [color={rgb, 255:red, 184; green, 233; blue, 134 }  ,draw opacity=1 ][line width=3]  (358,64) .. controls (358,50.19) and (369.19,39) .. (383,39) .. controls (396.81,39) and (408,50.19) .. (408,64) .. controls (408,77.81) and (396.81,89) .. (383,89) .. controls (369.19,89) and (358,77.81) .. (358,64) -- cycle ;
\draw  [color={rgb, 255:red, 80; green, 227; blue, 194 }  ,draw opacity=1 ][line width=3.75]  (418,233) .. controls (418,219.19) and (429.19,208) .. (443,208) .. controls (456.81,208) and (468,219.19) .. (468,233) .. controls (468,246.81) and (456.81,258) .. (443,258) .. controls (429.19,258) and (418,246.81) .. (418,233) -- cycle ;

\draw (311,58) node  [align=left] {$v_1$};
\draw (456,55) node  [align=left] {$v_2$};
\draw (277,150) node  [align=left] {$r_1$};
\draw (390,150) node  [align=left] {$r_2$};
\draw (500,190) node  [align=left] {$t_1$};
\draw (217,71) node  [align=left] {$t_2$};
\draw (360,240) node  [align=left] {$g_1$};
\draw (208,191) node  [align=left] {$g_2$};
\draw (269,231) node  [align=left] {$w_3$};
\draw (443,233) node  [align=left] {$h_{15}$};
\draw (165,129) node  [align=left] {$h_{16}$};
\draw (383,64) node  [align=left] {$l_1$};
\draw (154,52) node  [align=left] {$l_2$};
\end{tikzpicture}
    \caption{A schematic example of the graph representation.}
    \label{fig:graph}
\end{figure}
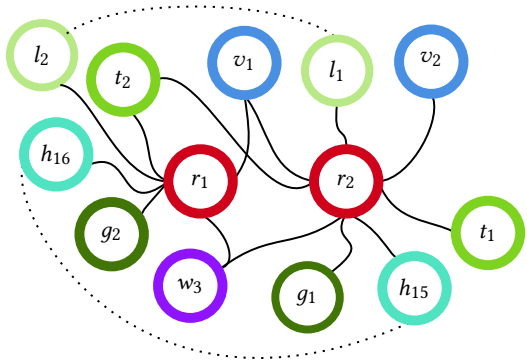
The best performing textual classifier has been implemented in a system that routes citizen reported urban micro events to the correct department of the City of Amsterdam. This allowed for the evaluation of this classifier in a real world setting and to gather multimodal data for the multimodal experiment.

The task of the implemented classifier is identifying the correct class using textual data. Whenever a prediction is made with a high enough probability the classifier automatically routes the customer service requests to the corresponding department. When the probability of the prediction is not higher than the threshold experts will perform the task of the classification. 
After the classification has been made, by the classifier or by the expert, it is possible that a mistake is made. Whenever a department receives a customer service request that is not classified correctly they will correct it to the correct class. This allows evaluation of both the classifier and experts performing the same task. Also customer evaluation is done by asking the initial reporter to rate the entire process.

The implemented textual classifier was used to gather data of performance in a real world setting, the resultst can be seen in Table \ref{table:customersat}. The textual classifier scored 89\% accuracy in a real world setting, experts performing the same task using all modalities for the classification score 91\% accuracy.

When looking at a survey done for 2768 customers reports, we find that reports that have been automatically classified receive a customer satisfaction of 3,2/5. The manually classified reports receive a 2,9/5. 
This leads us to believe that a decrease in time needed to resolve the issue likely had a positive effect on customer satisfaction.
Since the textual classifier is outperformed by an expert that can see location, time and the added image as well it is likely performance can increase by adding more modalities to the classifier. Improved classification performance in combination with much lower processing time than in case of manual annotation seems to be a promising path to an increased customer satisfaction.

\subsection{Multimodal Fusion}
\label{section:multimodal}
To asses what kind of fusion method works best for the multimodal classification of urban micro-events we will consider a number of different methods, starting with a simple early fusion, followed by the creation of a graph embedding. After this we will discuss a late fusion and a hybrid fusion method of creating a classifier.

The classifications for all types of fusion are done with XGBoost and a Logistic Regression to be able to evaluate performance of the classifiers on different modalities and fusion techniques. 

\subsubsection{Early fusion}

For every possible combination of feature sets a classifier will be created. 

Here the features are simply concatenated together and used as input for the classifier. An example of a early fusion is: 

$Visual \| Textual \| Geo \| Time \| Weather$

\begin{figure*}[h]
\centering
\includegraphics[width=1\textwidth]{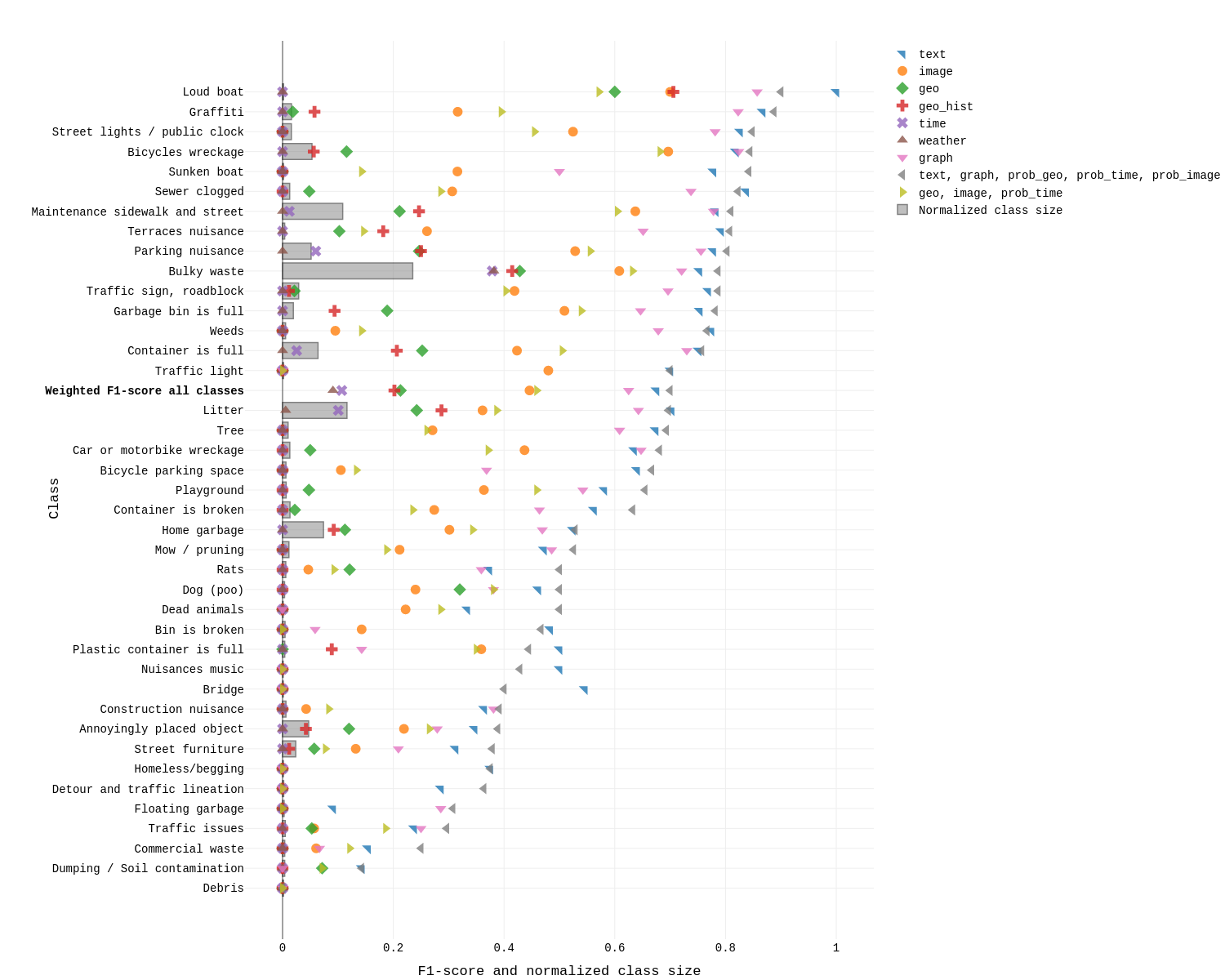}
\caption{Performance of issue level classes $(n=57)$ for all no fusion classifiers and the best performing fusion, with and without text.
Logistic regression was used for classification. F1-score used for evaluation. Normalized support added for reference. The 40 classes with the highest support.}
\label{fig:graph_sub_hybrid}
\end{figure*}

\subsubsection{Graph embedding}

To represent different modalities in one representation it is possible to place them in a graph. To evaluate this method a graph has been created for classification purposes. Let $G = (V, E)$ be our undirected weighted graph with the set of nodes $V$ and the set of edges $E$. We choose to use an undirected graph since the relations between the nodes and edges are symmetric.

The following nodes are added:

\begin{itemize}
    \item Report nodes $R = \{r_1,r_2, ..., r_n\}$ For every report in the dataset a node is created.
    \item Geo objects.  $G = \{g_1,g_2, ..., g_n\}$ For all reports the closest geo objects are added.
    \item Geo location.  $L = \{l_1,l_2, ..., l_n\}$ For all report the coordinates are added as a node.
    \item Visual concepts.  $V = \{v_1,v_2, ..., v_n\}$ For all images the top two visual concepts are added to the graph.
    \item Words. $T = \{t_1,t_2, ..., t_n\}$ The top 5000 words in the corpus are added to the graph.
    \item Time. $H = \{h_1,h_2, ..., h_{24}\}$ and $W = \{w_1,w_2, ..., w_7\}$ For every weekday and hour a node is created.
\end{itemize}

And the following edges will be created:
\begin{itemize}
    \item From the reports edges are created to the two closest geo objects, the weight of the edge is the distance.
    \item The reports are linked to the top two visual objects with the highest probability, with that probability as the weight of the edge.
    \item For all the words in the text of the report an edge is created, with TF-IDF as weight of the edge.
    \item For the weekday and hour of the report a binary weighted edge is added, and for all the neighboring weekdays and hours also and edge is created.
    \item Also an edge from the report to the geographical location is made. The geographical location is linked to the top two closest geographical location.
\end{itemize}

In Figure \ref{fig:graph} a schematic overview is given of the connection between edges and nodes. After the creation of the graph node2vec \cite{GroverL16} was used to create a 256-dimensional representation for every node. The node2vec frameworks learns low-dimensional representations for all the nodes in the graph. This is done by optimizing a neighborhood preserving objective by simulating random walks. The reports representation will then be used as a feature for classification and evaluation purposes.

\subsubsection{Late fusion}

Stacking will be used as ensemble learning technique to combine information out of two or more models into a new model. This is done by using the probabilistic output of classifiers as input for a new classifier. An example of a late fusion is: 

$Prob Visual \| Prob Textual \| Prob Geo \| Prob Time \| Prob Weather$

\subsubsection{Hybrid}

The hybrid fusion classifier is a combination of the early fusion and the late fusion classifier. For some feature sets the late fusion probabilistic output is used. For other feature sets the original features will be as in the early fusion. These features are combined as input for the classifier. For every possible combinations of early and late fusion features a classifier will be created and the predictions are evaluated on the test set to determine what the optimal method of hybrid fusion is on this specific dataset. An example of a hybrid fusion is: 

$Text \| Prob Visual \| Prob Geo \| Prob Time \| Prob Weather$

\section{Experimental setup}
\label{section:experimental_setup}
In this section we describe the evaluation criteria and the data we use for the experiments.

\subsection{Evaluation criteria}
For evaluation weighted F1 is used since the smaller classes are also important to classify correctly and we seek for a balance of recall and precision. Evaluation will be done with the same train/test split 80/20 for all experiments to allow for comparison.

\subsection{Data}
The data used is a set of citizen reports from the City of Amsterdam.
For different experiments multiple different subsets have been used:

\begin{itemize}
    \item 523.651 reports with textual information and their corresponding issue class have been used for creating a textual classifier.
    \item Of those reports 29.408 reports also had visual data.
    \item 9.362 of the reports with visual data had their label corrected by a domain expert.
\end{itemize}

The multimodal experiments will be done on the subset of data of 29.408 reports that has visual data. The reports can be grouped in eight main issue classes. The largest group is garbage, most of these reports are about bulky waste. Other issues in this group are litter, full garbage containers, broken garbage containers and construction waste. The second largest group is about anything related to roads, traffic and furniture. This group consists of issues about the maintenance of roads, traffic signs, clogging drains, slippery roads, broken streetlights, issues with playgrounds and dangerous traffic situations. The third largest group is about disturbance in the public space, ranging from bicycle wracks, illegal parking, objects blocking the sidewalk, noise nuisance and dog poo. The group of green and water is about any green that needs maintenance or quay wall that needs to be repaired. Animals is about disturbances from rats, wasps or pigeons. Disturbances by people, business or boats consists of noise nuisance, smell disturbances or e.g. speeding boats. There is noise in the labels since they are collected from a real world system, making it possible an urban micro-event is labelled incorrectly.

\begin{table}
 \caption{Customer satisfaction and performance (Accuracy with eight classes) of domain experts and a textual classifier.}
  \begin{tabular}{lll}
    \toprule
    &Customer& Resolved in same\\
    &satisfaction& issue class\\
    \midrule
    Domain expert & 2.9/5 & 91\%\\
    Textual classifier & 3.2/5 & 89\%\\
  \bottomrule
\end{tabular}

   \label{table:customersat}
\end{table}

\begin{table}[]
\caption{Top textual and non textual results for best hybrid fusion experiments and no fusion. Main class $(n=8)$. Logistic regression (LR) and XGBoost (XGB) used for classification. Weighted F1-score is used for evaluation. }
\begin{tabular}{lll}
\toprule
Classifier & Features                                               & F1    \\
\midrule
\textbf{LR  }       & \textbf{text, graph, prob\_time, prob\_image }                  &\textbf{0.882} \\
XGB    & geo, image, graph, geo\_hist, prob\_text               & 0.875 \\
\textbf{LR }        & \textbf{text  }                                                 & \textbf{0.865} \\
XGB     & text                                                   & 0.851 \\
LR         & graph                                                  & 0.844 \\
XGB         & graph                                                  & 0.810 \\
XGB         & prob\_image, prob\_geo\_hist, & 0.737 \\
           & prob\_weather, time, geo  &       \\
LR         & image, geo\_hist, prob\_geo                            & 0.730 \\
LR         & image                                                  & 0.710 \\
XGB        & image                                                  & 0.706 \\
XGB        & geo\_hist                                              & 0.508 \\
XGB        & geo                                                    & 0.505 \\
LR         & geo                                                    & 0.481 \\
LR         & geo\_hist                                              & 0.464 \\
XGB        & weather                                                & 0.400 \\
XGB        & time                                                   & 0.391 \\
LR         & time                                                   & 0.380 \\
LR         & weather                                                & 0.380 \\
\bottomrule
\end{tabular}

\label{table:results_main}%
\end{table}

\begin{table}
  \caption{Evaluation of several textual classifying methods trained on 418.920 samples and evaluated on 104.731 samples. The task was detecting which of seven main classes the report belongs to.}

  \begin{tabular}{lll}
    \toprule
Classifier                                                         & F1 macro avg & F1 micro avg \\
    \midrule
TF-IDF + LR                                     & 0.79         & 0.87         \\
Bidirectional CNN+LSTM                                             & 0.72         & 0.88         \\
W2V on reports + LR             & 0.73         & 0.83         \\
W2V 160 combined \cite{lin2017sentinlp} + LR & 0.60         & 0.72        \\
  \bottomrule
  
\end{tabular}
  \label{tab:textual_classifier}
\end{table}

\section{Experimental results}
\label{section:experimental_results}

\begin{table}[]
\caption{Top textual and non textual results for best hybrid fusion experiments and no fusion. Issue level $(n=57)$. Logistic regression (LR) and XGBoost (XGB) used for classification. Weighted F1-score is used for evaluation.}
\begin{tabular}{lll}
\toprule
Classifier & Features                                             & F1    \\
\midrule
\textbf{LR}        & \textbf{text, graph, prob\_geo,}       & \textbf{0.700} \\
           & \textbf{prob\_time, prob\_image}      & \\
XGB        & prob\_text, prob\_time, image                        & 0.699 \\
\textbf{LR}         & \textbf{text  }                                               & \textbf{0.675} \\
XGB        & text                                                 & 0.655 \\
LR         & graph                                                & 0.625 \\
XGB        & graph                                                & 0.573 \\
XGB        & geo\_hist, prob\_time, geo,   & 0.474 \\
           & prob\_image, weather ... &  \\
LR         & image                                                & 0.446 \\
XGB        & image                                                & 0.427 \\
XGB        & geo\_hist                                            & 0.258 \\
XGB        & geo                                                  & 0.255 \\
LR         & geo                                                  & 0.213 \\
LR         & geo\_hist                                            & 0.202 \\
XGB        & weather                                              & 0.129 \\
XGB        & time                                                 & 0.120 \\
LR         & time                                                 & 0.107 \\
LR         & weather                                              & 0.091 \\
\bottomrule
\end{tabular}

\label{table:results_sub}%
\end{table}

\begin{table*}[!htb]
\caption{Several examples of classification with example features.}
\begin{tabular}{llll}
\toprule
 & Example 1 & Example 2 & Example 3\\
\midrule
Image & \includegraphics[width=0.25\textwidth,height=0.14\textwidth]{bike.jpg} & \includegraphics[width=0.25\textwidth,height=0.14\textwidth]{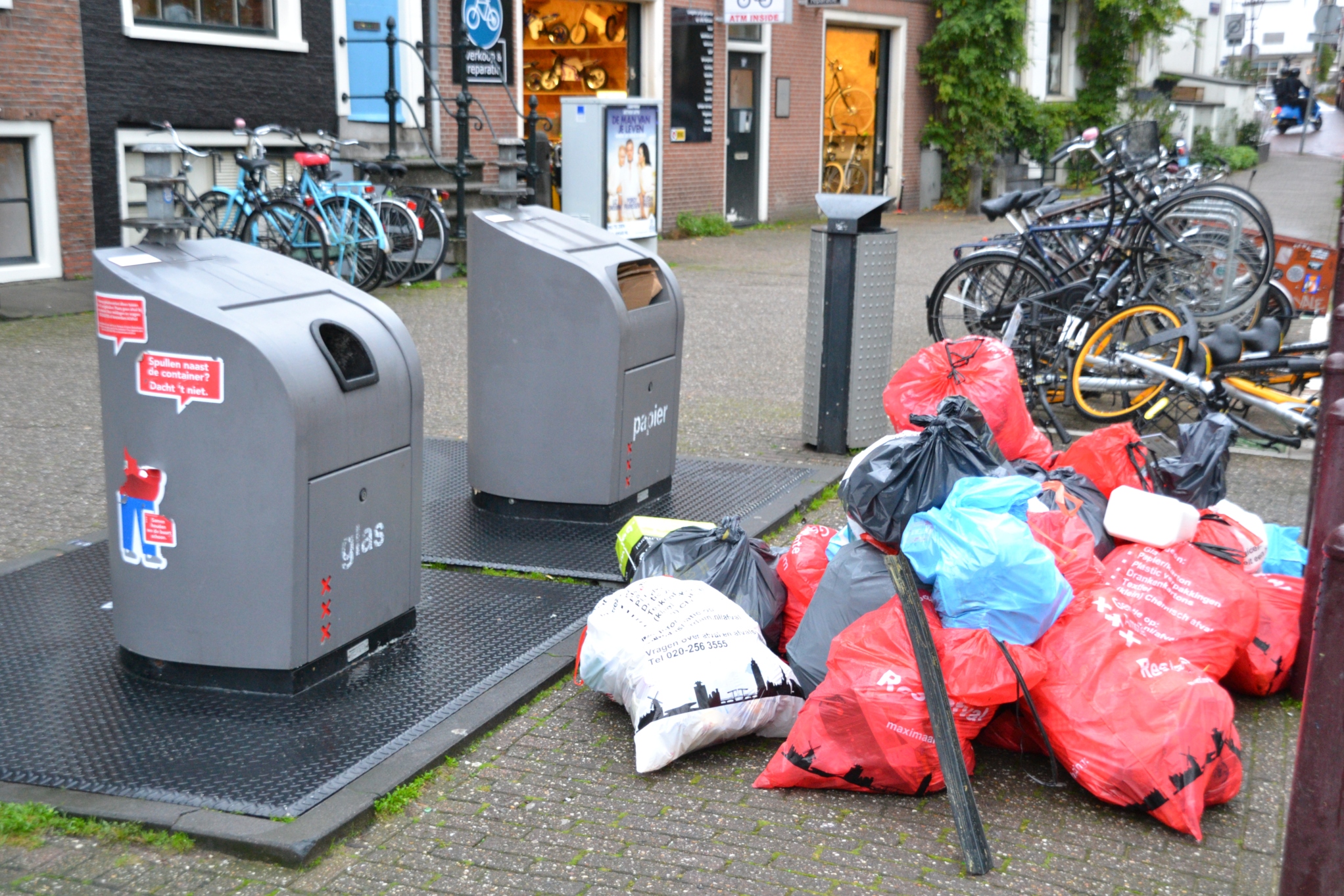} & \includegraphics[width=0.25\textwidth,height=0.14\textwidth]{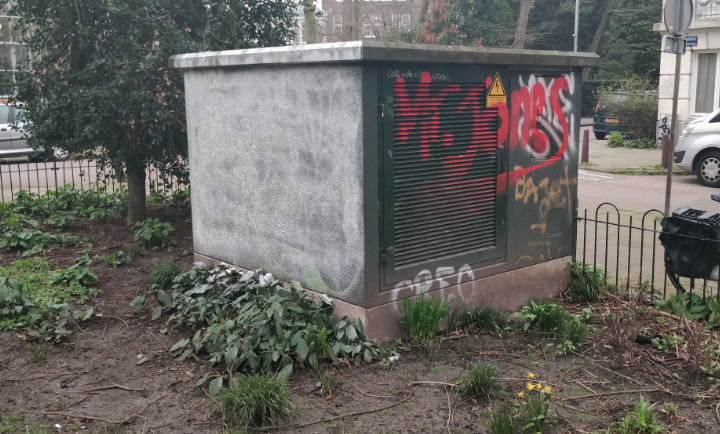}\\
\midrule
Visual features & mountain\_bike = 0.81 & ash\_can = 0.62 & chainlink\_fence = 0.17\\
 &bicycle-built-for-two = 0.03 & turnstile = 0.02 & ashcan = 0.10\\
 \midrule
Text                               &  They are still here & There is garbage next to the container & Graffiti\\
\midrule
Geo                              &  bicycles\_wreck\_within\_100m = 6 & garbage\_container\_within\_200m = 7  & hist\_graffiti\_within\_100m = 3\\
&  bicycles\_wreck\_mean\_5\_closest = 0.04 km &hist\_garbage\_nearest = 0.01 km & hist\_graffiti\_nearest = 0.01 km\\
\midrule
Time          &  2018-12-25 \ 09.05  & 2018-11-22 15.30& 2018-12-22 15.39 \\
\midrule
Label                             &  \textcolor[rgb]{0.25,0.46,0.02}{Bicycles wreckage } & \textcolor[rgb]{1,0,0}{Bulky waste} & \textcolor[rgb]{0.25,0.46,0.02}{Graffiti }\\
\midrule
Prediction                               &  \textcolor[rgb]{0.25,0.46,0.02}{Bicycles wreckage } & \textcolor[rgb]{1,0,0}{Litter} & \textcolor[rgb]{0.25,0.46,0.02}{Graffiti  }\\
\midrule
True class                          &  Bicycles wreckage& Home garbage& Graffiti \\
\bottomrule
\end{tabular}

\label{table:examples}%
\end{table*}

In this section several experiments and their results are discussed, and we answer the following questions.
\begin{itemize}
 \item What method of textual classification works best for the classification of urban micro-events?
 \item What visual, geo, time and textual features can be used for the detection of urban micro-events?
 \item What type of fusion is effective for the classification of urban micro-events?
\end{itemize}

The source code of experiments can be found in the linked repository \cite{github2019}. 

\subsection{Textual classifier}

The best performing model when evaluating using F1 macro was an TF-IDF with a logistic regression. CNN+LSTM performed best when looking at overall performance, but failed detecting some smaller, yet important classes. Evaluation is shown in Table \ref{tab:textual_classifier}.

\subsection{Multimodal classifier}
The best performing multimodal classifier was created by using a Logistic regression for classification and a hybrid fusion method, for both class level classification and issue level classification, the results can be seen in  Tables \ref{table:results_sub} and \ref{table:results_main}. The following features have been used by the best performing classifier: textual, late fusion geographical objects, late fusion time, late fusion visual and the graph embedding. The unimodal classifiers using time, weather or geo features yielded a better performance when using XGBoost, but the multimodal classifiers and the unimodal classifiers using visual and text data performed better when using logistic regression.

The results show that it is possible to improve on the textual classifier. For the class level classification adding visual information, time information and the graph embedding to the text with a hybrid fusion method increased performance from .865 to .882. When comparing this increase with the experts annotations as seen in Table \ref{table:customersat}, it is possibility that on a larger dataset the multimodal classifier will outperform domain experts performing the same task while increasing customer satisfaction.

For the issue level classification adding visual, geo, time information and the graph embedding to the textual data allows for an increase in performance from 0.675 to 0.700. This is a small increase but it has to be taken into consideration that the data used partly has weak labels generated by using the textual data, this creates a bias for the performance of the textual classifier. Even with this bias the performance has increased, making the results promising. Also the data use has a large class unbalance as can be seen in Figure \ref{fig:graph_sub_hybrid}, making the size of the available training data a potential limitation. This makes it likely that when more data is available performance will increase even further. 

In Figure \ref{fig:graph_sub_hybrid} the f1-score can be seen per class and per feature set. For most classes the multimodal classifier improves on performance over the textual baseline. An example of classes that improve from the multimodal classifier are "Sunken boats", "Floating garbage" and "Dead animals". These are all classes that have a clear visual hint, like a boat, water or an animal on the image.

A few classes do not get classified better when using the multimodal classifier, examples of these classes are  "Loud Boats", "Nuisance from music" and "plastic container being full." A few example classification are shown in Table \ref{table:examples}.  The Load boats textual classifier scores perfectly in evaluation, and the  addition of other modalities decreases performance. For almost all classes with sufficient class size the multimodal classifier works better for than the textual classifier.

\section{Conclusion}
In this paper we investigated the potential for automatically classifying urban micro events based on heterogeneous information describing them in citizen reports, which ranges from text and image to metadata about event geolocation, time and weather. We further deploy a number of approaches for fusing information extracted from different modalities, including traditional early and late fusion, but also novel representation learning on graphs and hybrid fusion. Finally, we investigate contribution of individual modalities to the overall classification performance. The experiments were conducted on a real-world dataset collected from a live citizen reporting system. Our main conclusion is that multimodal classifier yields a higher performance than the unimodal alternatives. Text appears to be a single most important modality, which is expected since the citizens are usually careful when describing the issues. In addition, our experiments show that the representation learning on graphs is effective in embedding heterogeneous information extracted from all different modalities into a compact, but discriminative representation. Indeed, a hybrid fusion of such created representation with different modalities associated with the citizen reports emerges as the overall best performing classification approach. In our future work we will further investigate an undoubtedly large potential of multimodal graph embeddings and the possibilities of incorporating different fusion mechanisms into representation learning.


\FloatBarrier
\pagebreak
%
\bibliographystyle{ACM-Reference-Format}
\bibliography{lib}


\begin{thebibliography}{42}


\ifx \showCODEN    \undefined \def \showCODEN     #1{\unskip}     \fi
\ifx \showDOI      \undefined \def \showDOI       #1{#1}\fi
\ifx \showISBNx    \undefined \def \showISBNx     #1{\unskip}     \fi
\ifx \showISBNxiii \undefined \def \showISBNxiii  #1{\unskip}     \fi
\ifx \showISSN     \undefined \def \showISSN      #1{\unskip}     \fi
\ifx \showLCCN     \undefined \def \showLCCN      #1{\unskip}     \fi
\ifx \shownote     \undefined \def \shownote      #1{#1}          \fi
\ifx \showarticletitle \undefined \def \showarticletitle #1{#1}   \fi
\ifx \showURL      \undefined \def \showURL       {\relax}        \fi
\providecommand\bibfield[2]{#2}
\providecommand\bibinfo[2]{#2}
\providecommand\natexlab[1]{#1}
\providecommand\showeprint[2][]{arXiv:#2}

\bibitem[\protect\citeauthoryear{Awad, Butt, Curtis, Lee, Fiscus, Godil, Joy,
  Delgado, Smeaton, Graham, Kraaij, Qu\'{e}not, Magalhaes, Semedo, and
  Blasi}{Awad et~al\mbox{.}}{2018}]%
        {2018trecvidawad}
\bibfield{author}{\bibinfo{person}{George Awad}, \bibinfo{person}{Asad Butt},
  \bibinfo{person}{Keith Curtis}, \bibinfo{person}{Yooyoung Lee},
  \bibinfo{person}{Jonathan Fiscus}, \bibinfo{person}{Afzal Godil},
  \bibinfo{person}{David Joy}, \bibinfo{person}{Andrew Delgado},
  \bibinfo{person}{Alan~F. Smeaton}, \bibinfo{person}{Yvette Graham},
  \bibinfo{person}{Wessel Kraaij}, \bibinfo{person}{Georges Qu\'{e}not},
  \bibinfo{person}{Joao Magalhaes}, \bibinfo{person}{David Semedo}, {and}
  \bibinfo{person}{Saverio Blasi}.} \bibinfo{year}{2018}\natexlab{}.
\newblock \showarticletitle{TRECVID 2018: Benchmarking Video Activity
  Detection, Video Captioning and Matching, Video Storytelling Linking and
  Video Search}. In \bibinfo{booktitle}{\emph{Proceedings of TRECVID 2018}}.
  NIST, USA.
\newblock


\bibitem[\protect\citeauthoryear{Bischke, Helber, Schulze, Srinivasan, Dengel,
  and Borth}{Bischke et~al\mbox{.}}{2017}]%
        {bischke2017multimedia}
\bibfield{author}{\bibinfo{person}{Benjamin Bischke}, \bibinfo{person}{Patrick
  Helber}, \bibinfo{person}{Christian Schulze}, \bibinfo{person}{Venkat
  Srinivasan}, \bibinfo{person}{Andreas Dengel}, {and} \bibinfo{person}{Damian
  Borth}.} \bibinfo{year}{2017}\natexlab{}.
\newblock \showarticletitle{The Multimedia Satellite Task at MediaEval 2017.}.
  In \bibinfo{booktitle}{\emph{MediaEval}}.
\newblock


\bibitem[\protect\citeauthoryear{Blei}{Blei}{2012}]%
        {Blei:2012:PTM:2133806.2133826}
\bibfield{author}{\bibinfo{person}{David~M. Blei}.}
  \bibinfo{year}{2012}\natexlab{}.
\newblock \showarticletitle{Probabilistic Topic Models}.
\newblock \bibinfo{journal}{\emph{Commun. ACM}} \bibinfo{volume}{55},
  \bibinfo{number}{4} (\bibinfo{date}{April} \bibinfo{year}{2012}),
  \bibinfo{pages}{77--84}.
\newblock
\showISSN{0001-0782}
\urldef\tempurl%
\url{https://doi.org/10.1145/2133806.2133826}
\showDOI{\tempurl}


\bibitem[\protect\citeauthoryear{Boididou, Papadopoulos, Kompatsiaris,
  Schifferes, and Newman}{Boididou et~al\mbox{.}}{2014}]%
        {boididou2014challenges}
\bibfield{author}{\bibinfo{person}{Christina Boididou}, \bibinfo{person}{Symeon
  Papadopoulos}, \bibinfo{person}{Yiannis Kompatsiaris}, \bibinfo{person}{Steve
  Schifferes}, {and} \bibinfo{person}{Nic Newman}.}
  \bibinfo{year}{2014}\natexlab{}.
\newblock \showarticletitle{Challenges of computational verification in social
  multimedia}. In \bibinfo{booktitle}{\emph{Proceedings of the 23rd
  International Conference on World Wide Web}}. ACM, \bibinfo{pages}{743--748}.
\newblock


\bibitem[\protect\citeauthoryear{Boulos, Resch, Crowley, Breslin, Sohn,
  Burtner, Pike, Jezierski, and Chuang}{Boulos et~al\mbox{.}}{2011}]%
        {boulos2011crowdsourcing}
\bibfield{author}{\bibinfo{person}{Maged N~Kamel Boulos},
  \bibinfo{person}{Bernd Resch}, \bibinfo{person}{David~N Crowley},
  \bibinfo{person}{John~G Breslin}, \bibinfo{person}{Gunho Sohn},
  \bibinfo{person}{Russ Burtner}, \bibinfo{person}{William~A Pike},
  \bibinfo{person}{Eduardo Jezierski}, {and} \bibinfo{person}{Kuo-Yu~Slayer
  Chuang}.} \bibinfo{year}{2011}\natexlab{}.
\newblock \showarticletitle{Crowdsourcing, citizen sensing and sensor web
  technologies for public and environmental health surveillance and crisis
  management: trends, OGC standards and application examples}.
\newblock \bibinfo{journal}{\emph{International journal of health geographics}}
  \bibinfo{volume}{10}, \bibinfo{number}{1} (\bibinfo{year}{2011}),
  \bibinfo{pages}{67}.
\newblock


\bibitem[\protect\citeauthoryear{Chen, Jakubowicz, Yang, Zhang, and Pan}{Chen
  et~al\mbox{.}}{2017}]%
        {chen2017fine}
\bibfield{author}{\bibinfo{person}{Longbiao Chen},
  \bibinfo{person}{J{\'e}r{\'e}mie Jakubowicz}, \bibinfo{person}{Dingqi Yang},
  \bibinfo{person}{Daqing Zhang}, {and} \bibinfo{person}{Gang Pan}.}
  \bibinfo{year}{2017}\natexlab{}.
\newblock \showarticletitle{Fine-Grained Urban Event Detection and
  Characterization Based on Tensor Cofactorization.}
\newblock \bibinfo{journal}{\emph{IEEE Trans. Human-Machine Systems}}
  \bibinfo{volume}{47}, \bibinfo{number}{3} (\bibinfo{year}{2017}),
  \bibinfo{pages}{380--391}.
\newblock


\bibitem[\protect\citeauthoryear{Chen, He, and Kan}{Chen et~al\mbox{.}}{2016}]%
        {chen2016context}
\bibfield{author}{\bibinfo{person}{Tao Chen}, \bibinfo{person}{Xiangnan He},
  {and} \bibinfo{person}{Min-Yen Kan}.} \bibinfo{year}{2016}\natexlab{}.
\newblock \showarticletitle{Context-aware image tweet modelling and
  recommendation}. In \bibinfo{booktitle}{\emph{Proceedings of the 24th ACM
  international conference on Multimedia}}. ACM, \bibinfo{pages}{1018--1027}.
\newblock


\bibitem[\protect\citeauthoryear{Collell, Zhang, and Moens}{Collell
  et~al\mbox{.}}{2017}]%
        {collell2017imagined}
\bibfield{author}{\bibinfo{person}{Guillem Collell}, \bibinfo{person}{Ted
  Zhang}, {and} \bibinfo{person}{Marie-Francine Moens}.}
  \bibinfo{year}{2017}\natexlab{}.
\newblock \showarticletitle{Imagined Visual Representations as Multimodal
  Embeddings.}. In \bibinfo{booktitle}{\emph{AAAI}}.
  \bibinfo{pages}{4378--4384}.
\newblock


\bibitem[\protect\citeauthoryear{Deng, Dong, Socher, Li, Li, and Fei-Fei}{Deng
  et~al\mbox{.}}{2009}]%
        {deng2009imagenet}
\bibfield{author}{\bibinfo{person}{Jia Deng}, \bibinfo{person}{Wei Dong},
  \bibinfo{person}{Richard Socher}, \bibinfo{person}{Li-Jia Li},
  \bibinfo{person}{Kai Li}, {and} \bibinfo{person}{Li Fei-Fei}.}
  \bibinfo{year}{2009}\natexlab{}.
\newblock \showarticletitle{Imagenet: A large-scale hierarchical image
  database}. In \bibinfo{booktitle}{\emph{2009 IEEE conference on computer
  vision and pattern recognition}}. Ieee, \bibinfo{pages}{248--255}.
\newblock


\bibitem[\protect\citeauthoryear{Grover and Leskovec}{Grover and
  Leskovec}{2016}]%
        {GroverL16}
\bibfield{author}{\bibinfo{person}{Aditya Grover} {and} \bibinfo{person}{Jure
  Leskovec}.} \bibinfo{year}{2016}\natexlab{}.
\newblock \showarticletitle{node2vec: Scalable Feature Learning for Networks}.
\newblock \bibinfo{journal}{\emph{CoRR}}  \bibinfo{volume}{abs/1607.00653}
  (\bibinfo{year}{2016}).
\newblock
\showeprint[arxiv]{1607.00653}
\urldef\tempurl%
\url{http://arxiv.org/abs/1607.00653}
\showURL{%
\tempurl}


\bibitem[\protect\citeauthoryear{He, Zhang, Ren, and Sun}{He
  et~al\mbox{.}}{2016}]%
        {he2016deep}
\bibfield{author}{\bibinfo{person}{Kaiming He}, \bibinfo{person}{Xiangyu
  Zhang}, \bibinfo{person}{Shaoqing Ren}, {and} \bibinfo{person}{Jian Sun}.}
  \bibinfo{year}{2016}\natexlab{}.
\newblock \showarticletitle{Deep residual learning for image recognition}. In
  \bibinfo{booktitle}{\emph{Proceedings of the IEEE conference on computer
  vision and pattern recognition}}. \bibinfo{pages}{770--778}.
\newblock


\bibitem[\protect\citeauthoryear{Hu}{Hu}{2018}]%
        {hu2018decode}
\bibfield{author}{\bibinfo{person}{Tianran Hu}.}
  \bibinfo{year}{2018}\natexlab{}.
\newblock \showarticletitle{Decode Human Life from Social Media}. In
  \bibinfo{booktitle}{\emph{2018 ACM Multimedia Conference}}. ACM,
  \bibinfo{pages}{820--824}.
\newblock


\bibitem[\protect\citeauthoryear{Huang, Li, and Shan}{Huang
  et~al\mbox{.}}{2018}]%
        {huang2018spatial}
\bibfield{author}{\bibinfo{person}{Yuqian Huang}, \bibinfo{person}{Yue Li},
  {and} \bibinfo{person}{Jie Shan}.} \bibinfo{year}{2018}\natexlab{}.
\newblock \showarticletitle{Spatial-Temporal Event Detection from Geo-Tagged
  Tweets}.
\newblock \bibinfo{journal}{\emph{ISPRS International Journal of
  Geo-Information}} \bibinfo{volume}{7}, \bibinfo{number}{4}
  (\bibinfo{year}{2018}), \bibinfo{pages}{150}.
\newblock


\bibitem[\protect\citeauthoryear{Jayarajah and Misra}{Jayarajah and
  Misra}{2016}]%
        {jayarajah2016can}
\bibfield{author}{\bibinfo{person}{Kasthuri Jayarajah} {and}
  \bibinfo{person}{Archan Misra}.} \bibinfo{year}{2016}\natexlab{}.
\newblock \showarticletitle{Can instagram posts help characterize urban
  micro-events?}. In \bibinfo{booktitle}{\emph{Information Fusion (FUSION),
  2016 19th International Conference on}}. IEEE, \bibinfo{pages}{130--137}.
\newblock


\bibitem[\protect\citeauthoryear{Jin, Cao, Guo, Zhang, and Luo}{Jin
  et~al\mbox{.}}{2017}]%
        {jin2017multimodal}
\bibfield{author}{\bibinfo{person}{Zhiwei Jin}, \bibinfo{person}{Juan Cao},
  \bibinfo{person}{Han Guo}, \bibinfo{person}{Yongdong Zhang}, {and}
  \bibinfo{person}{Jiebo Luo}.} \bibinfo{year}{2017}\natexlab{}.
\newblock \showarticletitle{Multimodal fusion with recurrent neural networks
  for rumor detection on microblogs}. In \bibinfo{booktitle}{\emph{Proceedings
  of the 25th ACM international conference on Multimedia}}. ACM,
  \bibinfo{pages}{795--816}.
\newblock


\bibitem[\protect\citeauthoryear{Kanojia, Kumar, and Ramamritham}{Kanojia
  et~al\mbox{.}}{2016}]%
        {kanojia2016civique}
\bibfield{author}{\bibinfo{person}{Diptesh Kanojia},
  \bibinfo{person}{Vishwajeet Kumar}, {and} \bibinfo{person}{Krithi
  Ramamritham}.} \bibinfo{year}{2016}\natexlab{}.
\newblock \showarticletitle{Civique: Using Social Media to Detect Urban
  Emergencies}.
\newblock \bibinfo{journal}{\emph{arXiv preprint arXiv:1610.04377}}
  (\bibinfo{year}{2016}).
\newblock


\bibitem[\protect\citeauthoryear{Kiela and Bottou}{Kiela and Bottou}{2014}]%
        {kiela2014learning}
\bibfield{author}{\bibinfo{person}{Douwe Kiela} {and} \bibinfo{person}{L{\'e}on
  Bottou}.} \bibinfo{year}{2014}\natexlab{}.
\newblock \showarticletitle{Learning image embeddings using convolutional
  neural networks for improved multi-modal semantics}. In
  \bibinfo{booktitle}{\emph{Proceedings of the 2014 Conference on Empirical
  Methods in Natural Language Processing (EMNLP)}}. \bibinfo{pages}{36--45}.
\newblock


\bibitem[\protect\citeauthoryear{Kittler, Hater, and Duin}{Kittler
  et~al\mbox{.}}{1996}]%
        {kittler1996combining}
\bibfield{author}{\bibinfo{person}{Josef Kittler}, \bibinfo{person}{Mohamad
  Hater}, {and} \bibinfo{person}{Robert~PW Duin}.}
  \bibinfo{year}{1996}\natexlab{}.
\newblock \showarticletitle{Combining classifiers}. In
  \bibinfo{booktitle}{\emph{Proceedings of 13th international conference on
  pattern recognition}}, Vol.~\bibinfo{volume}{2}. IEEE,
  \bibinfo{pages}{897--901}.
\newblock


\bibitem[\protect\citeauthoryear{Li, Ellis, Ji, and Chang}{Li
  et~al\mbox{.}}{2016}]%
        {li2016event}
\bibfield{author}{\bibinfo{person}{Hongzhi Li}, \bibinfo{person}{Joseph~G
  Ellis}, \bibinfo{person}{Heng Ji}, {and} \bibinfo{person}{Shih-Fu Chang}.}
  \bibinfo{year}{2016}\natexlab{}.
\newblock \showarticletitle{Event specific multimodal pattern mining for
  knowledge base construction}. In \bibinfo{booktitle}{\emph{Proceedings of the
  24th ACM international conference on Multimedia}}. ACM,
  \bibinfo{pages}{821--830}.
\newblock


\bibitem[\protect\citeauthoryear{Li, Zhu, Huang, Lu, and Zhao}{Li
  et~al\mbox{.}}{2018}]%
        {li2018read}
\bibfield{author}{\bibinfo{person}{Jingjing Li}, \bibinfo{person}{Lei Zhu},
  \bibinfo{person}{Zi Huang}, \bibinfo{person}{Ke Lu}, {and}
  \bibinfo{person}{Jidong Zhao}.} \bibinfo{year}{2018}\natexlab{}.
\newblock \showarticletitle{I read, I saw, I tell: Texts Assisted Fine-Grained
  Visual Classification}. In \bibinfo{booktitle}{\emph{2018 ACM Multimedia
  Conference}}. ACM, \bibinfo{pages}{663--671}.
\newblock


\bibitem[\protect\citeauthoryear{Lin, Xie, Yu, and Lai}{Lin
  et~al\mbox{.}}{2017}]%
        {lin2017sentinlp}
\bibfield{author}{\bibinfo{person}{Shuying Lin}, \bibinfo{person}{Huosheng
  Xie}, \bibinfo{person}{Liang-Chih Yu}, {and} \bibinfo{person}{K~Robert Lai}.}
  \bibinfo{year}{2017}\natexlab{}.
\newblock \showarticletitle{SentiNLP at IJCNLP-2017 Task 4: Customer Feedback
  Analysis Using a Bi-LSTM-CNN Model}.
\newblock \bibinfo{journal}{\emph{Proceedings of the IJCNLP 2017, Shared
  Tasks}} (\bibinfo{year}{2017}), \bibinfo{pages}{149--154}.
\newblock


\bibitem[\protect\citeauthoryear{Liu, Groves, Hayakawa, Poncelas, and Liu}{Liu
  et~al\mbox{.}}{2018}]%
        {liu2018understanding}
\bibfield{author}{\bibinfo{person}{Chao-Hong Liu}, \bibinfo{person}{Declan
  Groves}, \bibinfo{person}{Akira Hayakawa}, \bibinfo{person}{Alberto
  Poncelas}, {and} \bibinfo{person}{Qun Liu}.} \bibinfo{year}{2018}\natexlab{}.
\newblock \showarticletitle{Understanding Meanings in Multilingual Customer
  Feedback}.
\newblock \bibinfo{journal}{\emph{arXiv preprint arXiv:1806.01694}}
  (\bibinfo{year}{2018}).
\newblock


\bibitem[\protect\citeauthoryear{Mehrotra, Sanner, Buntine, and Xie}{Mehrotra
  et~al\mbox{.}}{2013}]%
        {mehrotra2013improving}
\bibfield{author}{\bibinfo{person}{Rishabh Mehrotra}, \bibinfo{person}{Scott
  Sanner}, \bibinfo{person}{Wray Buntine}, {and} \bibinfo{person}{Lexing Xie}.}
  \bibinfo{year}{2013}\natexlab{}.
\newblock \showarticletitle{Improving lda topic models for microblogs via tweet
  pooling and automatic labeling}. In \bibinfo{booktitle}{\emph{Proceedings of
  the 36th international ACM SIGIR conference on Research and development in
  information retrieval}}. ACM, \bibinfo{pages}{889--892}.
\newblock


\bibitem[\protect\citeauthoryear{Mikolov, Chen, Corrado, and Dean}{Mikolov
  et~al\mbox{.}}{2013}]%
        {mikolov2013efficient}
\bibfield{author}{\bibinfo{person}{Tomas Mikolov}, \bibinfo{person}{Kai Chen},
  \bibinfo{person}{Greg Corrado}, {and} \bibinfo{person}{Jeffrey Dean}.}
  \bibinfo{year}{2013}\natexlab{}.
\newblock \showarticletitle{Efficient estimation of word representations in
  vector space}.
\newblock \bibinfo{journal}{\emph{arXiv preprint arXiv:1301.3781}}
  (\bibinfo{year}{2013}).
\newblock


\bibitem[\protect\citeauthoryear{Mohandes, Deriche, and Aliyu}{Mohandes
  et~al\mbox{.}}{2018}]%
        {mohandes2018classifiers}
\bibfield{author}{\bibinfo{person}{Mohamed Mohandes}, \bibinfo{person}{Mohamed
  Deriche}, {and} \bibinfo{person}{Salihu~O Aliyu}.}
  \bibinfo{year}{2018}\natexlab{}.
\newblock \showarticletitle{Classifiers combination techniques: A comprehensive
  review}.
\newblock \bibinfo{journal}{\emph{IEEE Access}}  \bibinfo{volume}{6}
  (\bibinfo{year}{2018}), \bibinfo{pages}{19626--19639}.
\newblock


\bibitem[\protect\citeauthoryear{Offenhuber}{Offenhuber}{2014}]%
        {offenhuber2014infrastructure}
\bibfield{author}{\bibinfo{person}{Dietmar Offenhuber}.}
  \bibinfo{year}{2014}\natexlab{}.
\newblock \showarticletitle{Infrastructure legibility - a comparative analysis
  of open311-based citizen feedback systems}.
\newblock \bibinfo{journal}{\emph{Cambridge Journal of Regions, Economy and
  Society}} (\bibinfo{year}{2014}), \bibinfo{pages}{rsu001}.
\newblock


\bibitem[\protect\citeauthoryear{Park, Kim, Park, Cha, Nam, Yoon, and
  Rhim}{Park et~al\mbox{.}}{2015}]%
        {Park2015mining}
\bibfield{author}{\bibinfo{person}{Kunwoo Park}, \bibinfo{person}{Jaewoo Kim},
  \bibinfo{person}{Jaram Park}, \bibinfo{person}{Meeyoung Cha},
  \bibinfo{person}{Jiin Nam}, \bibinfo{person}{Seunghyun Yoon}, {and}
  \bibinfo{person}{Eunhee Rhim}.} \bibinfo{year}{2015}\natexlab{}.
\newblock \showarticletitle{Mining the Minds of Customers from Online Chat
  Logs}. In \bibinfo{booktitle}{\emph{Proceedings of the 24th ACM International
  on Conference on Information and Knowledge Management}}
  \emph{(\bibinfo{series}{CIKM '15})}. \bibinfo{publisher}{ACM},
  \bibinfo{address}{New York, NY, USA}, \bibinfo{pages}{1879--1882}.
\newblock
\showISBNx{978-1-4503-3794-6}
\urldef\tempurl%
\url{https://doi.org/10.1145/2806416.2806621}
\showDOI{\tempurl}


\bibitem[\protect\citeauthoryear{Perozzi, Al-Rfou, and Skiena}{Perozzi
  et~al\mbox{.}}{2014}]%
        {Perozzi:2014:DOL:2623330.2623732}
\bibfield{author}{\bibinfo{person}{Bryan Perozzi}, \bibinfo{person}{Rami
  Al-Rfou}, {and} \bibinfo{person}{Steven Skiena}.}
  \bibinfo{year}{2014}\natexlab{}.
\newblock \showarticletitle{DeepWalk: Online Learning of Social
  Representations}. In \bibinfo{booktitle}{\emph{Proceedings of the 20th ACM
  SIGKDD International Conference on Knowledge Discovery and Data Mining}}
  \emph{(\bibinfo{series}{KDD '14})}. \bibinfo{publisher}{ACM},
  \bibinfo{address}{New York, NY, USA}, \bibinfo{pages}{701--710}.
\newblock
\showISBNx{978-1-4503-2956-9}
\urldef\tempurl%
\url{https://doi.org/10.1145/2623330.2623732}
\showDOI{\tempurl}


\bibitem[\protect\citeauthoryear{Qian, Zhang, and Xu}{Qian
  et~al\mbox{.}}{2016}]%
        {qian2016multi}
\bibfield{author}{\bibinfo{person}{Shengsheng Qian}, \bibinfo{person}{Tianzhu
  Zhang}, {and} \bibinfo{person}{Changsheng Xu}.}
  \bibinfo{year}{2016}\natexlab{}.
\newblock \showarticletitle{Multi-modal multi-view topic-opinion mining for
  social event analysis}. In \bibinfo{booktitle}{\emph{Proceedings of the 24th
  ACM international conference on Multimedia}}. ACM, \bibinfo{pages}{2--11}.
\newblock


\bibitem[\protect\citeauthoryear{Ramos et~al\mbox{.}}{Ramos
  et~al\mbox{.}}{2003}]%
        {ramos2003using}
\bibfield{author}{\bibinfo{person}{Juan Ramos} {et~al\mbox{.}}}
  \bibinfo{year}{2003}\natexlab{}.
\newblock \showarticletitle{Using tf-idf to determine word relevance in
  document queries}. In \bibinfo{booktitle}{\emph{Proceedings of the first
  instructional conference on machine learning}}, Vol.~\bibinfo{volume}{242}.
  \bibinfo{pages}{133--142}.
\newblock


\bibitem[\protect\citeauthoryear{Rudinac, Chua, Diaz-Ferreyra, Friedland,
  Gornostaja, Huet, Kaptein, Lind{\'e}n, Moens, Peltonen,
  et~al\mbox{.}}{Rudinac et~al\mbox{.}}{2018}]%
        {rudinac2018rethinking}
\bibfield{author}{\bibinfo{person}{Stevan Rudinac}, \bibinfo{person}{Tat-Seng
  Chua}, \bibinfo{person}{Nicolas Diaz-Ferreyra}, \bibinfo{person}{Gerald
  Friedland}, \bibinfo{person}{Tatjana Gornostaja}, \bibinfo{person}{Benoit
  Huet}, \bibinfo{person}{Rianne Kaptein}, \bibinfo{person}{Krister
  Lind{\'e}n}, \bibinfo{person}{Marie-Francine Moens}, \bibinfo{person}{Jaakko
  Peltonen}, {et~al\mbox{.}}} \bibinfo{year}{2018}\natexlab{}.
\newblock \showarticletitle{Rethinking summarization and storytelling for
  modern social multimedia}. In \bibinfo{booktitle}{\emph{International
  Conference on Multimedia Modeling}}. Springer, \bibinfo{pages}{632--644}.
\newblock


\bibitem[\protect\citeauthoryear{Rudinac, Hanjalic, and Larson}{Rudinac
  et~al\mbox{.}}{2013}]%
        {rudinac2013generating}
\bibfield{author}{\bibinfo{person}{Stevan Rudinac}, \bibinfo{person}{Alan
  Hanjalic}, {and} \bibinfo{person}{Martha Larson}.}
  \bibinfo{year}{2013}\natexlab{}.
\newblock \showarticletitle{Generating visual summaries of geographic areas
  using community-contributed images}.
\newblock \bibinfo{journal}{\emph{IEEE Transactions on Multimedia}}
  \bibinfo{volume}{15}, \bibinfo{number}{4} (\bibinfo{year}{2013}),
  \bibinfo{pages}{921--932}.
\newblock


\bibitem[\protect\citeauthoryear{Snoek, Worring, and Smeulders}{Snoek
  et~al\mbox{.}}{2005}]%
        {snoek2005early}
\bibfield{author}{\bibinfo{person}{Cees~GM Snoek}, \bibinfo{person}{Marcel
  Worring}, {and} \bibinfo{person}{Arnold~WM Smeulders}.}
  \bibinfo{year}{2005}\natexlab{}.
\newblock \showarticletitle{Early versus late fusion in semantic video
  analysis}. In \bibinfo{booktitle}{\emph{Proceedings of the 13th annual ACM
  international conference on Multimedia}}. ACM, \bibinfo{pages}{399--402}.
\newblock


\bibitem[\protect\citeauthoryear{Sukel}{Sukel}{2019}]%
        {github2019}
\bibfield{author}{\bibinfo{person}{M. Sukel}.} \bibinfo{year}{2019}\natexlab{}.
\newblock \bibinfo{title}{Multimodal classification of urban micro-events}.
\newblock
  \bibinfo{howpublished}{\url{https://github.com/maartensukel/multimodal-classification-of-urban-micro-events}}.
\newblock


\bibitem[\protect\citeauthoryear{Tang, Pongpaichet, and Jain}{Tang
  et~al\mbox{.}}{2016a}]%
        {tang2016research}
\bibfield{author}{\bibinfo{person}{Mengfan Tang}, \bibinfo{person}{Siripen
  Pongpaichet}, {and} \bibinfo{person}{Ramesh Jain}.}
  \bibinfo{year}{2016}\natexlab{a}.
\newblock \showarticletitle{Research challenges in developing multimedia
  systems for managing emergency situations}. In
  \bibinfo{booktitle}{\emph{Proceedings of the 24th ACM international
  conference on Multimedia}}. ACM, \bibinfo{pages}{938--947}.
\newblock


\bibitem[\protect\citeauthoryear{Tang, Pongpaichet, and Jain}{Tang
  et~al\mbox{.}}{2016b}]%
        {Tang:2016:RCD:2964284.2976761}
\bibfield{author}{\bibinfo{person}{Mengfan Tang}, \bibinfo{person}{Siripen
  Pongpaichet}, {and} \bibinfo{person}{Ramesh Jain}.}
  \bibinfo{year}{2016}\natexlab{b}.
\newblock \showarticletitle{Research Challenges in Developing Multimedia
  Systems for Managing Emergency Situations}. In
  \bibinfo{booktitle}{\emph{Proceedings of the 24th ACM International
  Conference on Multimedia}} \emph{(\bibinfo{series}{MM '16})}.
  \bibinfo{publisher}{ACM}, \bibinfo{address}{New York, NY, USA},
  \bibinfo{pages}{938--947}.
\newblock
\showISBNx{978-1-4503-3603-1}
\urldef\tempurl%
\url{https://doi.org/10.1145/2964284.2976761}
\showDOI{\tempurl}


\bibitem[\protect\citeauthoryear{Van~Gysel, de~Rijke, and Worring}{Van~Gysel
  et~al\mbox{.}}{2016}]%
        {VanGysel:2016:UES:2872427.2882974}
\bibfield{author}{\bibinfo{person}{Christophe Van~Gysel},
  \bibinfo{person}{Maarten de Rijke}, {and} \bibinfo{person}{Marcel Worring}.}
  \bibinfo{year}{2016}\natexlab{}.
\newblock \showarticletitle{Unsupervised, Efficient and Semantic Expertise
  Retrieval}. In \bibinfo{booktitle}{\emph{Proceedings of the 25th
  International Conference on World Wide Web}} \emph{(\bibinfo{series}{WWW
  '16})}. \bibinfo{publisher}{International World Wide Web Conferences Steering
  Committee}, \bibinfo{address}{Republic and Canton of Geneva, Switzerland},
  \bibinfo{pages}{1069--1079}.
\newblock
\showISBNx{978-1-4503-4143-1}
\urldef\tempurl%
\url{https://doi.org/10.1145/2872427.2882974}
\showDOI{\tempurl}


\bibitem[\protect\citeauthoryear{Walther and Kaisser}{Walther and
  Kaisser}{2013}]%
        {walther2013geo}
\bibfield{author}{\bibinfo{person}{Maximilian Walther} {and}
  \bibinfo{person}{Michael Kaisser}.} \bibinfo{year}{2013}\natexlab{}.
\newblock \showarticletitle{Geo-spatial event detection in the twitter stream}.
  In \bibinfo{booktitle}{\emph{European Conference on Information Retrieval}}.
  Springer, \bibinfo{pages}{356--367}.
\newblock


\bibitem[\protect\citeauthoryear{Wang, Yang, Xu, Hanjalic, and Shen}{Wang
  et~al\mbox{.}}{2017}]%
        {wang2017adversarial}
\bibfield{author}{\bibinfo{person}{Bokun Wang}, \bibinfo{person}{Yang Yang},
  \bibinfo{person}{Xing Xu}, \bibinfo{person}{Alan Hanjalic}, {and}
  \bibinfo{person}{Heng~Tao Shen}.} \bibinfo{year}{2017}\natexlab{}.
\newblock \showarticletitle{Adversarial cross-modal retrieval}. In
  \bibinfo{booktitle}{\emph{Proceedings of the 25th ACM international
  conference on Multimedia}}. ACM, \bibinfo{pages}{154--162}.
\newblock


\bibitem[\protect\citeauthoryear{Yang, Zhang, and Xu}{Yang
  et~al\mbox{.}}{2016}]%
        {yang2016abnormal}
\bibfield{author}{\bibinfo{person}{Xiaoshan Yang}, \bibinfo{person}{Tianzhu
  Zhang}, {and} \bibinfo{person}{Changsheng Xu}.}
  \bibinfo{year}{2016}\natexlab{}.
\newblock \showarticletitle{Abnormal event discovery in user generated photos}.
  In \bibinfo{booktitle}{\emph{Proceedings of the 24th ACM international
  conference on Multimedia}}. ACM, \bibinfo{pages}{47--51}.
\newblock


\bibitem[\protect\citeauthoryear{Yang, Pierce, and Carbonell}{Yang
  et~al\mbox{.}}{1998}]%
        {yang1998study}
\bibfield{author}{\bibinfo{person}{Yiming Yang}, \bibinfo{person}{Tom Pierce},
  {and} \bibinfo{person}{Jaime Carbonell}.} \bibinfo{year}{1998}\natexlab{}.
\newblock \showarticletitle{A study of retrospective and on-line event
  detection}. In \bibinfo{booktitle}{\emph{Proceedings of the 21st annual
  international ACM SIGIR conference on Research and development in information
  retrieval}}. ACM, \bibinfo{pages}{28--36}.
\newblock


\bibitem[\protect\citeauthoryear{Zhang, Whitehead, Zhang, Li, Ellis, Huang,
  Liu, Ji, and Chang}{Zhang et~al\mbox{.}}{2017}]%
        {zhang2017improving}
\bibfield{author}{\bibinfo{person}{Tongtao Zhang}, \bibinfo{person}{Spencer
  Whitehead}, \bibinfo{person}{Hanwang Zhang}, \bibinfo{person}{Hongzhi Li},
  \bibinfo{person}{Joseph Ellis}, \bibinfo{person}{Lifu Huang},
  \bibinfo{person}{Wei Liu}, \bibinfo{person}{Heng Ji}, {and}
  \bibinfo{person}{Shih-Fu Chang}.} \bibinfo{year}{2017}\natexlab{}.
\newblock \showarticletitle{Improving event extraction via multimodal
  integration}. In \bibinfo{booktitle}{\emph{Proceedings of the 25th ACM
  international conference on Multimedia}}. ACM, \bibinfo{pages}{270--278}.
\newblock


\end{thebibliography}

%
\appendix

\end{document}